\documentclass[letterpaper, 10 pt, journal, twoside]{IEEEtran}
\usepackage{svg}
\usepackage{graphics} % for pdf, bitmapped graphics files
\usepackage{amsmath} % assumes amsmath package installed
\usepackage{amssymb}  % assumes amsmath package installed
\usepackage[font=footnotesize,labelformat=simple]{subcaption}
\usepackage{tabularray}
\usepackage{multirow}
\usepackage{array}
\usepackage{booktabs}
\newcolumntype{N}{>{\centering\arraybackslash}m{.2in}}
\newcolumntype{G}{>{\centering\arraybackslash}m{.2in}}
\usepackage{colortbl}
\usepackage{cite}

\begin{document}

\title{A Modular Residual Learning Framework to Enhance Model-Based Approach for Robust Locomotion}

\author{Min-Gyu Kim$^{1}$, Dongyun Kang$^{1}$, Hajun Kim$^{1}$, and Hae-Won Park$^{1}$%
\thanks{Manuscript received: March, 31, 2025; Revised June, 21, 2025; Accepted July, 13, 2025.}%Use only for final RAL version
\thanks{This paper was recommended for publication by Editor Abderrahmane Kheddar upon evaluation of the Associate Editor and Reviewers' comments.
This work was supported by the Technology Innovation Program(or Industrial Strategic Technology Development Program-Robot Industry Technology Development)(RS-2024-00427719, Dexterous and Agile Humanoid Robots for Industrial Applications) funded By the Ministry of  Trade Industry \& Energy(MOTIE, Korea); in part by Korea Evaluation Institute of Industrial Technology (KEIT) funded by the Korea Government (MOTIE) under Grant No.20018216, Development of mobile intelligence SW for autonomous navigation of legged robots in dynamic and atypical environments for real application.} %Use only for final RAL version
\thanks{$^{1}$The authors are with Humanoid Robot Research Center, Department of Mechanical Engineering, Korea Advanced Institute of Science and Technology (KAIST), Yuseong-gu 34141 Daejeon, Republic of Korea.
        {\tt\footnotesize haewonpark@kaist.ac.kr}}%
\thanks{Digital Object Identifier (DOI): see top of this page.}
}

\markboth{IEEE Robotics and Automation Letters. Preprint Version. Accepted July, 2025}
{Kim \MakeLowercase{\textit{et al.}}: A Modular Residual Learning Framework to Enhance Model-Based Approach for Robust Locomotion}


\maketitle

\begin{abstract}
This paper presents a novel approach that combines the advantages of both model-based and learning-based frameworks to achieve robust locomotion. The residual modules are integrated with each corresponding part of the model-based framework, a footstep planner and dynamic model designed using heuristics, to complement performance degradation caused by a model mismatch. By utilizing a modular structure and selecting the appropriate learning-based method for each residual module, our framework demonstrates improved control performance in environments with high uncertainty, while also achieving higher learning efficiency compared to baseline methods.
Moreover, we observed that our proposed methodology not only enhances control performance but also provides additional benefits, such as making nominal controllers more robust to parameter tuning. To investigate the feasibility of our framework, we demonstrated residual modules combined with model predictive control in a real quadrupedal robot. Despite uncertainties beyond the simulation, the robot successfully maintains balance and tracks the commanded velocity.
\end{abstract}


\IEEEpeerreviewmaketitle

\section{Introduction}

% \IEEEPARstart{T}{his} is the first sentence of my Introduction.
\IEEEPARstart{L}{egged} systems have fascinated researchers due to their potential to navigate urban and harsh environments while effectively performing assigned tasks. 
Model-based approaches (MBA), particularly model predictive control (MPC), have been widely studied for their capacity to handle system dynamics and constraints. 
Despite the rise of learning-based approaches (LBA), MBA remains central to generate safe and consistent motion.

However, MBA faces challenges, most notably model mismatches resulting from necessary simplifications for real-time feasibility. 
Since legged systems exhibit complex hybrid dynamics due to contact, the models used for control design are typically simplified, such as the single rigid body model or the linear inverted pendulum model.
While this reduces computational cost, it can lead to the loss of critical information such as contact dynamics, degrading performance in unmodeled scenarios.

To address these limitations, LBA can offer a promising complement, excelling in modeling complex behaviors through data. A synergistic integration of MBA and LBA can combine the reliability of MBA with the adaptability of LBA, yielding robust control even under significant uncertainties.

\begin{figure}[t]
    \centering
    \includegraphics[width=0.99\linewidth]{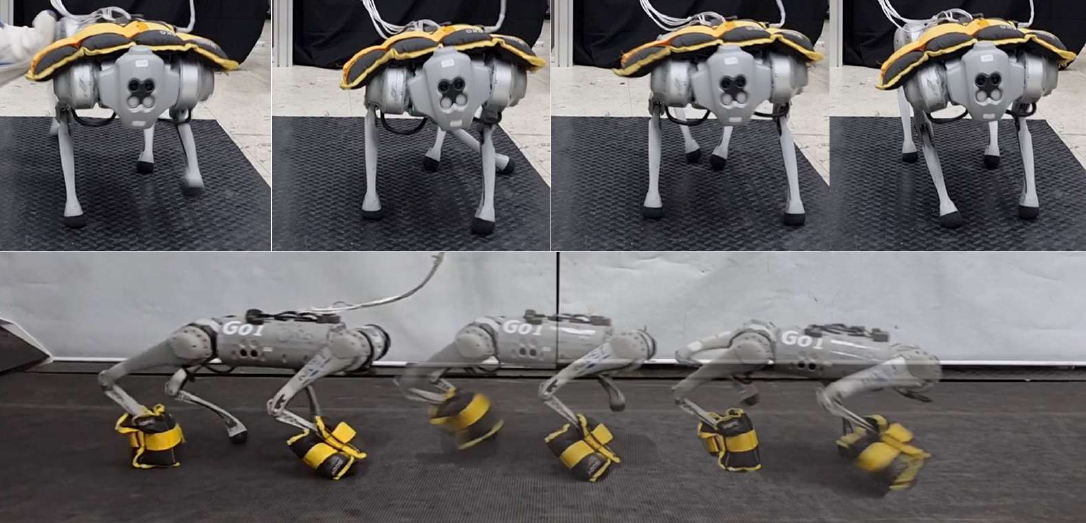}
    \caption{Snapshots of experimental results. The quadrupedal robot with conventional model-based controller adjusted by residual modules can overcome uncertainties such as unknown payload and disturbances.}
    \label{fig:00}
\end{figure}

This paper presents a hybrid scheme that leverages LBA to compensate for limitations in existing MBA.
While MBA offers refined and reliable motion, they are prone to degraded performance due to minor model inaccuracies or suboptimal heuristics, as well as significant uncertainties such as heavy payload or external disturbances.

To tackle this issue, we introduce \textit{residual modules} into two key components, the footstep planner and the dynamics model. 
Each residual module is designed using machine learning techniques to provide auxiliary actions. 
As a result, the hybrid controller offers improved robustness to uncertainty than a conventional MBA, while also preserving consistency outside the training domain compared to the end-to-end LBA. 
Fig. \ref{fig:00} shows an example of robustness, where a robot handles a payload and disturbance that the nominal MBA alone cannot cope with.

A key advantage of our method is its modular and simplified module design using reinforcement learning (RL) and supervised learning (SL) rather than fully relying on RL, which enables efficient learning while maintaining the nominal controller in the training loop. 
Specifically, we propose two residual modules: (i) a RL-based residual footstep module to adaptively correct foothold patterns in response to disturbances, and (ii) a SL-based residual dynamics (RD) module to account for discrepancies in the system model. 

We address challenges of foothold selection from complex contact dynamics via RL, while compensating continuous-domain dynamics discrepancies through SL, separating them from RL training.
This approach reduces the learning search space and simplifies the RL process.
To further reduce computational overhead, we employ convex MPC with a simplified dynamics model for 3D motion. 
Additionally, we apply low-pass filtering to the RD to isolate slowly varying uncertainties, thereby enhancing robustness to noise. By focusing on these slow variations, we assume the residual terms remain constant over the MPC prediction horizon, which further simplifies the optimization problem.
Compared to baselines, our method stands out by its architectural simplicity, improved training efficiency, and consistent performance across a wide range of out-of-distribution (OOD) scenarios. 
We highlight our contributions as follows.

\begin{itemize}
\item We propose a hybrid scheme that leverages learning-based residual modules to compensate for performance degradation caused by model inaccuracies and suboptimal heuristics in conventional MBA.
\item The framework features a lightweight modular architecture using RL for footstep adaptation and SL for continuous-domain dynamics correction, combined with simplified nominal dynamics and filtering–based RD decoupling to enable efficient learning.
\item Extensive experiments show that our method achieves robust and reliable task execution under heavy disturbances and OOD conditions, reduced parameter tuning sensitivity, and improved learning efficiency without compromising performance compared to baselines.
\end{itemize}

%%%%%%%%%%%%%%%%%%%%%%%%%%%%%%%%%%%%%%%%%%%%%%%%%%%%%%%%%%%%%%%%%%%%%%%%%%%%%%%%

\section{RELATED WORK} \label{ch:02}

\subsection{Model- and Learning-based Approaches} \label{ch:02-01}

Conventionally, heuristic-based schemes have been introduced for controlling legged robots, wherein the original system is represented by simplified models, such as a linear inverted pendulum model (LIPM) or single rigid body model (SRBM) \cite{raibert1986legged, pratt2006capture, di2018dynamic}. 
This scheme has been verified to be simple but practical through a range of demonstrations.

Among these, MPC has gained attention for its ability to handle dynamic tasks and system constraints \cite{wensing2023optimization, di2018dynamic, kim2019highly, hong2020real}.
However, its computational cost often necessitates simplifications, such as predefined gait sequences or simplified dynamics such as SRBM, leading to performance degradation.

In contrast, data-driven approaches have demonstrated impressive results in challenging tasks  \cite{ji2022concurrent, miki2022learning, li2023robust, cheng2024extreme}, but also face issues such as suboptimal convergence.
While techniques like system identification \cite{yu2017preparingunknownlearninguniversal, kumar2021rmarapidmotoradaptation} and online adaptation \cite{9001157, smith2022walkparklearningwalk} improve flexibility, end-to-end RL methods still struggle with issues such as sample inefficiency and a cumbersome reward engineering process.

\subsection{Residual Estimation} \label{ch:02-02}
To address the performance degradation of MBA caused by model discrepancy, numerous residual estimation approaches have been proposed across various domains such as legged locomotion, drones, and vehicles. 
These methods vary based on how accurately the residuals represent the actual discrepancies and whether the residual is updated online or designed offline for use in a hierarchical manner.

Specifically, residual estimation techniques range from adaptive control \cite{minniti2021adaptive} to data-based approaches, including online regression using linear model \cite{9525285}, 
probabilistic regression based on basis function learning \cite{arcari2023bayesian} or Gaussian processes \cite{kabzan2019learning}, 
and offline-trained neural network models \cite{Bauersfeld__2021, 9691797, 10049101}. 
Each method exhibits a trade-off between accuracy, real-time capability, and ease of integration into optimization-based controllers, depending on its complexity.

While compensating for dynamics gaps via residual estimation is effective in handling various disturbances and external loads, it may be insufficient on its own to achieve robust control in unstructured environments, particularly for unstable and highly dynamic systems such as legged robots.

\subsection{Hybrid Method} \label{ch:02-03}

To overcome the limitations of MBA and LBA, recent studies explore hybrid strategies to merge their strengths. 
Some works use LBA to replace heuristically defined high-level references (e.g. commands or gait sequences) for MBA for better adaptability in uncertain environments \cite{10.1007/978-3-031-21090-7_31, pmlr-v164-yang22d, 9719129, pmlr-v211-yang23b, gangapurwala2022rloc}. 

Additional approaches apply residual action, such as torque or joint reference, to directly fine-tune the nominal output \cite{9104757, BELLEGARDA2024104799, 10610453, gangapurwala2022rloc}.
While this can improve reactivity by enabling residuals to override nominal degradation, it may pose instability with uncertainties due to the absence of constrained optimization-based feasibility checks in MBA.

Finally, several studies employ MBA to accelerate the initial learning phase of RL, leveraging their ability to generate refined motions \cite{10225268, 10268037}. 
Similarly, MBA capable of long-term planning, such as trajectory optimization, has been used to address challenges of RL in sparse reward learning \cite{jenelten2024dtc, fuchioka2023opt}. 
While these approaches improve learning efficiency and convergence, they also introduce computational overhead due to repeated MBA computation during training.

%%%%%%%%%%%%%%%%%%%%%%%%%%%%%%%%%%%%%%%%%%%%%%%%%%%%%%%%%%%%%%%%%%%%%%%%%%%%%%%%

\section{RESIDUAL MODULES WITH NOMINAL CONTROLLER} \label{ch:03}

\begin{figure*}[t]
  \centering
  \includegraphics[width=0.95\linewidth]{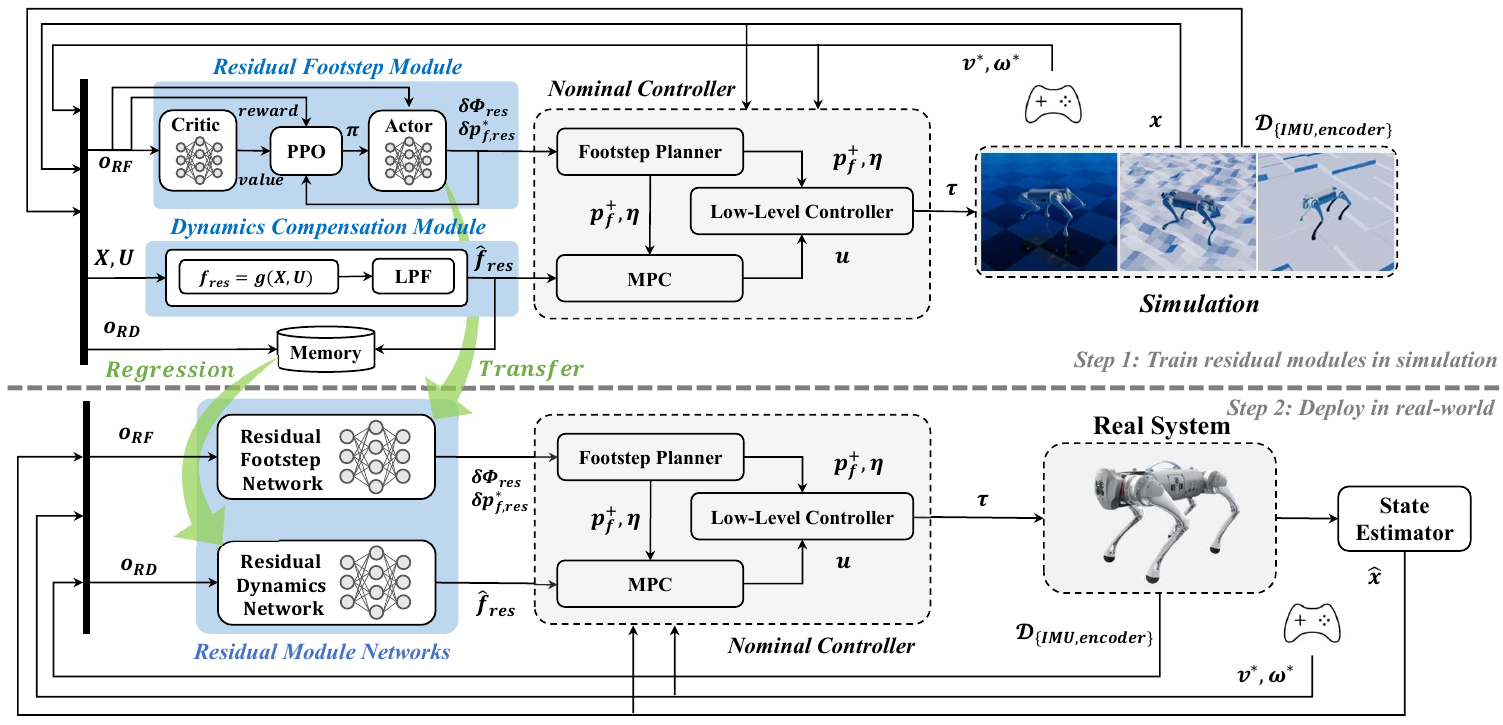}
  \caption{An illustration of the overall architecture. Each residual module finds auxiliary actions for compensating model mismatches in corresponding nominal modules. 
  The residual footstep module is learned using RL in simulation, while RD data is simultaneously collected and fed into the nominal controller during the learning process. The RD module is then reconstructed into a neural network for real-world deployment.
  }
  \label{fig:01}
\end{figure*}

\subsection{System Overview} \label{ch:03-01}

Our approach integrates learning-based residual modules into a MBA, as depicted in Fig \ref{fig:01}. Each module is designed to modulate the output of its corresponding nominal module, for improving task-specific performance. 

A key aspect of the proposed method is the sequential design of residual modules utilizing both RL and SL, depending on the characteristics of each component. 
In related work on drones, SL has been used to model aerodynamic or actuator dynamics from real-world data \cite{Bauersfeld__2021, 9691797, 10049101}. 
However, legged robots present additional challenges due to the need for choosing footholds.

To address this, we first introduce an RL-based residual module to compensate the heuristic footstep planner, enabling adaptive foothold adjustment. 
In parallel, we compute discrepancies in the nominal continuous dynamics analytically using simulation data. 
These discrepancies are then reconstructed through SL, using proprioceptive sensor data history, to ensure reliable performance in real-world.
This approach enhances sample efficiency by reducing the number of variables optimized during the RL process without compromising performance compared to other hybrid methods.

\subsection{Nominal Hybrid Dynamics} \label{ch:03-02}

This section explains a hybrid dynamics model of legged system and corresponding residual modules. The discrete hybrid dynamics can be expressed as follows \cite{1166523}.
\begin{gather}
    {\dot{\textbf{x}}} = \textbf{f}(\textbf{x},\textbf{u}),\quad \textbf{x}\notin S\\
    \textbf{x}^+ = \Delta(\textbf{x}^-),\quad \textbf{x}\in S \label{eq:discrete-dynamics}
\end{gather}
where $\textbf{x}$ is a state; 
$\textbf{u}$ is a ground reaction force (GRF); 
$S$ is a switching set which includes every moment of contact; 
$\textbf{f}(\textbf{x,u})$ is a continuous dynamics; 
and $\Delta(\textbf{x})$ is a switching dynamics.
Throughout this paper, vector-valued variables are denoted in boldface, while matrix-valued variables are represented using calligraphic font.
When contact states do not change, the system follows the continuous dynamics as follows.
\begin{gather}
    \textbf{x} = [\textbf{p},\ \boldsymbol{\phi},\ \dot{\textbf{p}},\ \boldsymbol{\omega}]\in\mathbb{R}^{12}, \\
    \dot{\textbf{x}} =\textbf{f}_n(\textbf{x,u})
    = \begin{bmatrix}
        \dot{\textbf{p}} \\
        \mathcal{M}(\boldsymbol{\phi}_0) \boldsymbol{\omega} \\
        \frac{1}{M}\sum^{4}_{i=1}{\textbf{u}_{i}} + \textbf{g} \\
        \mathcal{I}^{-1}_\mathbb{W}\sum^{4}_{i=1}{(\textbf{p}_{f,i}\times \textbf{u}_{i})}
    \end{bmatrix},
    \label{eq:nominal-dynamics}
\end{gather}
where $\textbf{f}_n$ is a nominal continuous dynamics; 
$\textbf{p}$ is a position of center of mass (CoM); 
$\textbf{p}_f$ is a foothold; 
$\boldsymbol{\phi}$ is Euler angle; 
$\boldsymbol{\omega}$ is angular velocity of CoM expressed in the body frame $\mathbb{B}$; 
$\mathcal{M}(\boldsymbol{\phi}_0)$ is a mapping matrix to transform angular velocity into Euler angle rate at operating point $\boldsymbol{\phi}_0$; 
$M$ is a lumped mass of the robot; 
$\textbf{g}$ is a gravity; 
$\mathcal{I}_\mathbb{W}$ is an inertia matrix in world frame $\mathbb{W}$ with a following relationship $\mathcal{I}_\mathbb{W}=\mathcal{R}(\boldsymbol{\phi}_0)\mathcal{I}_\mathbb{B}\mathcal{R}(\boldsymbol{\phi}_0)^T$; 
and $\mathcal{R}(\cdot)$ is a rotation matrix. 

In this work, we choose a SRBM-based convex MPC as a nominal controller in \cite{di2018dynamic}. This framework is well-established in many research and has some favorable features. For instance, the convex optimization formulation is advantageous for massive learning procedure by ensuring fast computation time by exploiting state-of-the-art solvers.

\subsection{Residual Footstep Module} \label{ch:03-03}

For simplification, we assume that footholds instantaneously change whenever contact state changes. Therefore, we hierarchically define the footstep variables in a separate heuristic-based planner as suggested in \cite{pratt2006capture}. The foothold and gait pattern of the $i$-th leg are defined as follows.
\begin{gather}
    \textbf{p}_{f,i}^+ = \textbf{p}_{f,i}^- + \delta \textbf{p}_{f,heuristic,i},\quad \textbf{x}\in S \\
    \Phi_{k,i} = \Phi_{k-1,i} + \frac{\delta t}{T_{step,i}},
\end{gather}
where $\delta \textbf{p}_{f,heuristic}$ are reference foothold from the heuristic planner; 
$\Phi\in[0,1]$ is a gait phase parameter; 
$\delta t$ is a control time step; 
and $T_{step}$ is a footstep time indicating swing and stance phases. 
For example, $\Phi$ for a leg initialized in stance starts at zero and increments by $\delta t/T_{stance}$ each control cycle. When $\Phi$ reaches 1, it is reset to zero, and $T_{step}$ is set to $T_{swing}$, repeating the same procedure.

Since the heuristic footstep planner comprises a fixed gait pattern and simplified model, which cannot fully reflect actual dynamics such as the inertial effect or non-flat terrain, the residual footstep module compensates it as follows.
\begin{gather}
    \textbf{p}_{f,i}^+ = \textbf{p}_{f,i}^- + \delta \textbf{p}_{f,heuristic,i} + \delta \textbf{p}_{f,res,i},\quad \textbf{x}\in S \\
    \Phi_{k,i} = \Phi_{k-1,i} + \frac{\delta t}{T_{step,i}} + \delta \Phi_{res,k-1,i},
\end{gather}
where $\delta \textbf{p}_{f,res},\ \delta \Phi_{res}$ are residual footstep and phase.

\begin{table}[t]
\caption{List of Domain Randomization}
\begin{center}
\begin{tabular}{|c|c|c|}
\hline
Parameter & Notation & Range \\ 
\hline
Command velocity & $[v^*_{x},v^*_{y},\omega^*_{z}]$ & $\pm[2.0, 1.0, 1.0]\ m(rad)/s$\\
\hline
Friction coefficient & $\mu$ & $[0.4,\ 1.0]$\\
\hline
Payload & $M_{payload}$ & $[-1.0,\ 5.0]\ kg$\\
\hline
Bumpiness & $h_{env}$ & $[0, 0.1]\ m$\\
\hline
CoM position & $\delta p_{CoM,xyz}$ & $\pm 0.05\ m$\\
\hline
Initial rotation & $\phi_{init}$ & $\pm 15\ deg$\\
\hline
Initial height$^*$ & $\delta z_{init}$ & $[0, 0.1]\ m$\\
\hline
\end{tabular}
\end{center}
\label{table:DR}
\caption*{\raggedleft\scriptsize *deviation from nominal height of CoM.}
\end{table}

The residual footstep module is learned using RL. 
The nominal controller employs a heuristic footstep planner that assumes flat terrain and no-slip conditions, which cannot fully handle more complex scenarios, such as rough or stepped terrain with varying friction coefficients. 
To address these limitations, we train the modules under three terrain types (flat, rough, stepped) with varying friction conditions.
Additionally, we randomize system and environmental parameters at each reset, as detailed in Table \ref{table:DR}, and initialize the robot at random body angles and heights slightly above the ground to further enhance robustness.

The reward function is designed to be similar to the cost function from MPC so that the module can operate as intended by nominal controller. 
Because the nominal controller can compute the walking pattern and required control inputs, it reduces the initial search space for learning, allowing the reward function to be simply designed as follows.
\begin{gather}
    r_{total} = r_{alive} + c_1||\textbf{x}^*-\textbf{x}|| + c_2||\boldsymbol{\tau}||, \\
    r_{alive} = \begin{cases}
        -10,\ \textit{if robot fails} \\
        0,\ \textit{otherwise}
    \end{cases}
\end{gather}
where $r_{alive}$ is a penalty whenever the system fails; 
$\textbf{x}^*$ is a user-defined desired state; 
$\boldsymbol{\tau}$ is a joint torque; 
and $c_i$ is a weight for each reward. 

The training is performed using proximal policy optimization (PPO) \cite{schulman2017proximalpolicyoptimizationalgorithms}, 
resulting in 16-dimensional output including residual foothold and phase of all legs. 
Details of the PPO setup are provided in Table \ref{table:PPO}.
The observation $\textbf{o}_{RF}$ for the residual footstep network $\boldsymbol{\pi}_{RF}$ is constructed as follows.
\begin{gather}
    [\delta \textbf{p}_{f,res,k}, \delta \Phi_{res,k}] = \boldsymbol{\pi}_{RF}(\textbf{o}_{RF,k}), \\
    \begin{split}
        \textbf{o}_{RF,k} =\ &[\dot{\textbf{p}}^*_k, \dot{\boldsymbol{\omega}}^*_k, \boldsymbol{\phi}_k, \boldsymbol{\omega}_k, 
        \{\boldsymbol{\theta},\dot{\boldsymbol{\theta}}\}_{j,\{k,\cdots,k-h\}},\\
         % \dot{\boldsymbol{\theta}}_{j,\{k,\cdots,k-h\}}, 
        &\delta \textbf{p}_{f,heuristic}, \Phi_{k,i}, \boldsymbol{\eta}, \boldsymbol{\tau}_k]\in\mathbb{R}^{113},
    \end{split}
\end{gather}
where $\dot{\textbf{p}}^*,\dot{\boldsymbol{\omega}}^*$ is a command velocity; 
$\boldsymbol{\theta}_{j,\{k,\cdots,k-h\}}$ is a history of joint positions with a window size $h$; 
$\boldsymbol{\eta}$ is a Boolean vector representing the planned contact sequence for each leg from the nominal gait planner; 
and $\boldsymbol{\tau}$ is a joint torque. 
Note that $h=2$ is used in this paper.

\subsection{Residual Dynamics Module} \label{ch:03-04}

The nominal model $\textbf{f}_n(\textbf{x},\textbf{u})$ in (\ref{eq:nominal-dynamics}) contains rotational dynamics which is locally time-invariant and linearized based on Euler angle. The state is then predicted using numerical forward Euler integration through the control horizon.
\begin{gather}
    \textbf{x}_{k+1} = \textbf{x}_k + \delta t\textbf{f}_n(\textbf{x}_k,\textbf{u}_k),
\end{gather}

In fact, this model does not fully account for actual whole-body dynamics or external disturbances, mainly due to the impact in switching phase, which can lead to performance degradation. This model discrepancy, $\textbf{f}_{res}$, can be determined using the history of states and control input as follows.
\begin{gather}
    \textbf{f}_{res,k-1} = \delta t^{-1}(\textbf{x}_{k}-\textbf{x}_{k-1}) - \textbf{f}_n(\textbf{x}_{k-1},\textbf{u}_{k-1}). 
    \label{eq:resdyn-analytic}
\end{gather}

\begin{table}[t]
\caption{PPO hyperparameter for Residual Footstep}
\begin{center}
\begin{tabular}{|c|c|}
\hline
Parameter & Value \\ 
\hline
Network size (actor \& critic) & $[256, 128]$\\
\hline
Activation function & LeakyReLU \\
\hline
\# of environments & $100$\\
\hline
\# of environment steps per update & $200$\\
\hline
\# of batches & $4$\\
\hline
\# of epochs & $4$\\
\hline
Learning rate & $0.0005$\\
\hline
Discount factor & $0.996$\\
\hline
GAE & $0.95$\\
\hline
Clip range & $0.2$\\
\hline
\end{tabular}
\end{center}
\label{table:PPO}
\end{table}

To deploy this term in real-world applications, several challenges must be addressed. 
First, since it is used directly as input to the MPC, excessive fluctuations in its value can significantly compromise the performance of the controller. 
Furthermore, while representing RD as a function of the MPC state and control input could enable more accurate dynamics prediction and potentially allow its use within the MPC optimization horizon \cite{10049101}, this approach would substantially increase computational burden for the RL process.

To mitigate these issues, we design a low-pass IIR filter:
\begin{gather}
    \hat{\textbf{f}}_{res,k} = e^{\frac{-2\pi F_c}{F_s}}\hat{\textbf{f}}_{res,k-1} + (1-e^{\frac{-2\pi F_c}{F_s}})\textbf{f}_{res,k},
    \label{eq:res-dyn}
\end{gather}
where $F_c,F_s$ are cutoff and sampling frequencies. We set $F_c=10\ Hz$, $F_s=1\ kHz$ in this work. 
The filter is designed to capture only the dominant low-frequency components of uncertainties, such as payloads or external disturbances, that can be considered quasi-static within each MPC control loop.
Additionally, it can prevent the system from being too sensitive to noisy signal. 

Nevertheless, the analytical RD in (\ref{eq:res-dyn}) still require accurate information, including linear velocity, contact states and GRF. 
Estimating this information in a real-world environment is challenging, and the resulting inaccurate measurement of RD can lead to catastrophic failure of the system.

A possible solution is to train a neural network after collecting data from simulations or experiments \cite{Bauersfeld__2021}.
We first collect simulation data to accurately label the RD while training the residual footstep module.
After training the footstep module, we reconstruct data set of RD and corresponding observations into an neural network model, $\boldsymbol{\pi}_{RD}$, to predict the residual term using only directly measurable proprioceptive sensor data, such as IMU and joint encoders. 
This model can then be trained using SL. 
\begin{gather}
    \hat{\textbf{f}}_{res,k} = \boldsymbol{\pi}_{RD}(\textbf{o}_{RD,k},\cdots,\textbf{o}_{RD,k-h}), \label{eq:resdyn} \\
    \textbf{o}_{RD,k} =\ [\boldsymbol{\phi}_k, \boldsymbol{\omega}_k, \boldsymbol{\theta}_{j,k}, \dot{\boldsymbol{\theta}}_{j,k}, \boldsymbol{\tau}_k, \hat{\textbf{f}}_{res,k-1}]\in\mathbb{R}^{54}.
\end{gather}

This network is designed with the same structure as the residual footstep module. We randomly collected a total of 10 million data to train this module.
This regression improves robustness in a real-world for inferring RD, compared to relying solely on an analytical design.

With aforementioned dynamics and footstep planning, the MPC tries to solve the following optimization problem to obtain control input.
\begin{align}
    & \underset{\textbf{X}, \textbf{U}}{\text{minimize}}
    & & \sum_{i=0}^{N-1}{l(\textbf{x}_{k+i},\textbf{u}_{k+i})} \label{eq:cost-function} \\
    & \text{subject to}
    & & \dot{\textbf{x}}_{k+i+1}=\textbf{f}_n(\textbf{x}_{k+i},\textbf{u}_{k+i})+\hat{\textbf{f}}_{res,k}, \\
    &&& \textbf{X}\in\mathbb{X},\ \textbf{U}\in\mathbb{U}
\end{align}
where $\textbf{X,U}$ is a set of states and control inputs through the horizon $N$; 
$l(\textbf{x},\textbf{u})$ is a cost function to minimize; 
and $\{\mathbb{X,U}\}$ is a feasible domain of state and control input. 
Note that the current residual term $\hat{\textbf{f}}_{res,k}$ is provided by the RD module and is independent of the state and control input. 
Consequently, it is applied uniformly across the entire prediction horizon at each control iteration. 
This filtering-based decoupling enables fast computation of MPC during both learning and real-world implementation.

The cost function is formulated as a least-square form.
\begin{gather}
    l(\textbf{X},\textbf{U}) = \textbf{X}^Tdiag(\textbf{w}_\textbf{x})\textbf{X} 
    + \textbf{U}^Tdiag(\textbf{w}_\textbf{u})\textbf{U}, \label{eq:least-square} \\ 
    \textbf{w}_\textbf{x} = \left[
    \textbf{w}_\textbf{p}, \textbf{w}_{\boldsymbol{\phi}}, \textbf{w}_\textbf{v}, \textbf{w}_{\boldsymbol{\omega}}
    \right],
\end{gather}
where $diag(\boldsymbol{\zeta})$ is a diagonal matrix consisting of diagonal vector $\boldsymbol{\zeta}$; 
and $\textbf{w}_{(\cdot)}$ is a weight for each variable. 

The feedforward and feedback joint torques are computed via leg kinematics as follows \cite{di2018dynamic}. 
\begin{gather}
    \boldsymbol{\tau}_{ff,i} = \mathcal{J}(q)_i^T\textbf{u}_i, \\ 
    \boldsymbol{\tau}_{fb,i} = \mathcal{J}(q)_i^T[\mathcal{K}_P(\textbf{p}_{f,i}^*-\textbf{p}_{f,i})
    +\mathcal{K}_D(\dot{\textbf{p}}_{f,i}^*-\dot{\textbf{p}}_{f,i})], \\
    \boldsymbol{\tau}_{i} = \boldsymbol{\tau}_{ff,i} + \boldsymbol{\tau}_{fb,i},
\end{gather}
where $\mathcal{J}_i$ is a foot Jacobian, $\textbf{p}_{f,i}^*$ is a reference foot position from footstep planner, and $\mathcal{K}_{P,D}$ are Cartesian PD gains.

%%%%%%%%%%%%%%%%%%%%%%%%%%%%%%%%%%%%%%%%%%%%%%%%%%%%%%%%%%%%%%%%%%%%%%%%%%%%%%%%
\section{RESULTS} \label{ch:04}

In this section, we conduct comparative experiments to evaluate the advantages of the proposed framework in terms of command tracking and robustness against uncertainties, compared to baseline controllers. 
These experiments assess the system’s ability to reliably handle a range of disturbances and model errors, such as kicks or heavy payloads.

Additionally, we examine whether the proposed method can reduce the control parameter dependency of the nominal MBA controller.
We also conduct OOD tests to evaluate whether our framework maintains consistent performance under conditions that differ significantly from those encountered during simulation, in comparison to existing end-to-end RL method. 
At last, we evaluate how our residual module design affects learning efficiency and convergence compared to baseline methods.

\subsection{Experimental Setup}

We validate our framework through simulations and real-world experiments using \textit{Unitree Go1}, a $12\ kg$ quadrupedal robot with 12 DoF. 
The simulation environment is built using the physics simulator \textit{RAISIM} \cite{raisim}. 
MPC and residual modules are updated at $100\ Hz$, while the remaining low-level parts run at $1\ kHz$. 
Note that the MPC predicts the next $0.1\ s$ with a time step of $0.01\ s$ over 10 horizons at each control iteration. 
We exploit a linear Kalman filter and momentum-based contact detection for the state estimation. 
The nominal stance and swing times are set to be $0.3\ s$, corresponding to a trot gait.  

\begin{table}[t]
\centering
\noindent
\caption{Experimental Result of Robustness Test}
{\scriptsize 
\begin{tabular}{N G G G}\toprule
\multicolumn{1}{c}{\textbf{Name}} &
    \multicolumn{1}{c}{\textbf{Normal (A)}} &
    \multicolumn{1}{c}{\textbf{With payload (B)}} &
    \multicolumn{1}{c}{\textbf{With disturbance (C)}} \\
\cmidrule(lr){2-4}
\multicolumn{1}{c}{\textit{vanilla-MPC}} &
    \multicolumn{1}{c}{0.0216} & 
    \multicolumn{1}{c}{\textit{Failed}} &
    \multicolumn{1}{c}{\textit{Failed}} \\
\multicolumn{1}{c}{\textit{res-dyn}} &
    \multicolumn{1}{c}{0.0118} & 
    \multicolumn{1}{c}{0.0538} &
    \multicolumn{1}{c}{\textit{Failed}} \\
\multicolumn{1}{c}{\textit{res-all}} &
    \multicolumn{1}{c}{0.0212} & 
    \multicolumn{1}{c}{0.0197} &
    \multicolumn{1}{c}{0.0695} \\
\bottomrule
\end{tabular}
\caption*{\raggedleft\scriptsize Values indicate RMS error of roll-angle.}
\label{table:robustness-test}
}
\end{table}
\begin{figure}[t]
    \centering
    \includegraphics[width=0.99\linewidth]{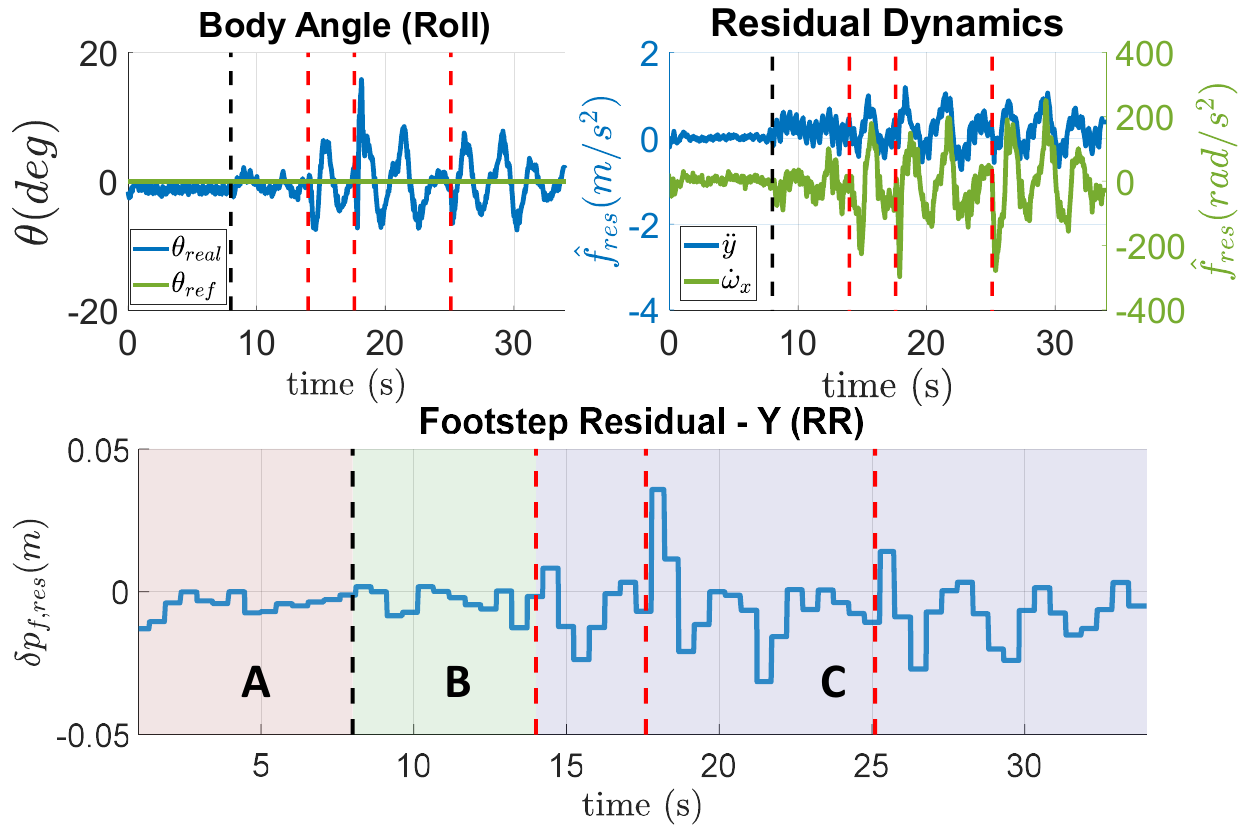}
    \caption{An experimental result of robustness test with \textit{res-all}. Each graph indicates the roll-angle (top-left), RD (top-right), and y-directional footstep residual of rear-right (RR) leg in the body frame (bottom). The black dotted line marks the moment when the payload is applied, while red lines indicate the disturbances.}
    \label{fig:03}
\end{figure}

\subsection{Effect of Residual Modules}

We first investigate the effects of each residual module. 
In this scenario, the robot is commanded to maintain its posture while external payload of  $6\ kg$ is exerted on its head with the following baselines: nominal MPC (\textit{vanilla-MPC}), MPC with only a RD module (\textit{res-dyn}), and MPC with both residual footstep and dynamics module (\textit{res-all}). 
Due to the payload being applied far from the CoM, a significant model error is expected. If the robot successfully adapts to the given payload, it is subjected to external disturbances.

\begin{figure}[t]
  \centering
  \includegraphics[width=0.99\linewidth]{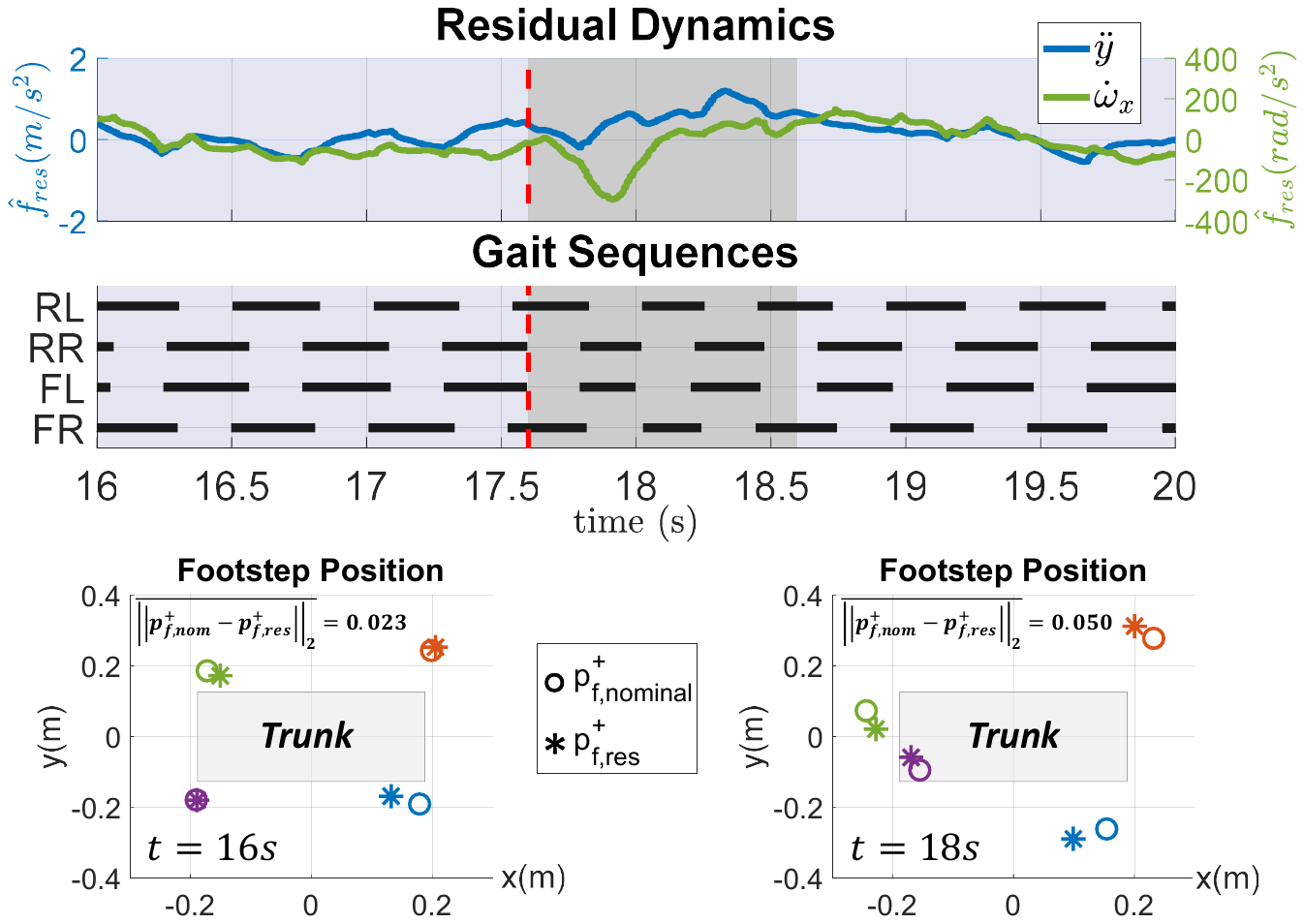}
  \caption{An detailed results of robustness test using \textit{res-all}. In the gait sequence plot, black lines indicate the stance phases of each corresponding leg, while the gray shaded area denotes the 1-second recovery period immediately following external disturbances applied to the robot.
  The footstep position graph shows the position of each footstep in a top-down view within the robot's body frame, where the gray rectangle in the graph represents the robot's trunk.
  }
  \label{fig:04}
\end{figure}

Table \ref{table:robustness-test} and Fig. \ref{fig:03} illustrate the results. 
The time intervals of each condition (normal, with payload, with disturbance) are denoted in Fig. \ref{fig:03}. 
As seen in Table \ref{table:robustness-test}, \textit{vanilla-MPC} fails to maintain its posture under the given payload. 

In contrast, \textit{res-dyn} allows the robot to compensate for the model discrepancy once the payload is applied. However, over time, the robot exhibits a sagittal swaying motion. Because the gait pattern remains fixed, it cannot adapt to the resulting motion, leading to failure when additional disturbances are exerted.

On the other hand, \textit{res-all} successfully controls locomotion under both the heavy payload and external disturbances.
As seen in Fig. \ref{fig:03}, the RD module captures model discrepancies whenever disturbances are exerted, while the residual footstep module modifies footholds to counteract the induced swaying motion.

Fig. \ref{fig:04} describes the detailed results from \textit{res-all}. Without external disturbances, the footholds and phase from the residual footstep module remain close to the heuristic. However, when a disturbance generates angular momentum in the roll direction, the module reduces stance time and adjusts footholds to stabilize the motion.

The proposed method is also evaluated under relatively high-speed locomotion, where the robot accelerated up to $1 m/s$ in the x-direction. In this experiment, a reduction in gait phase is observed during forward acceleration, as shown in the supplementary video.

\subsection{Robustness to Control Parameters}

It is widely known that the performance of MPC is highly sensitive to control parameters, such as weight in the cost function \cite{wensing2023optimization}. 
Despite their critical role, these parameters are difficult to optimize directly and are often selected empirically to simplify the problem \cite{9719129}. 
Similar to reward engineering in RL, this leads to a cumbersome tuning process. 
In this experiment, we investigate whether the residual modules can reduce this sensitivity, thereby improving robustness across a range of parameter configurations.

To evaluate this, the robot is tasked with velocity tracking on flat terrain under varying cost function weights. 
Specifically, during training, the weights in (\ref{eq:least-square}) are randomized within $\textbf{w}_{\textbf{p},\boldsymbol{\phi}}\in[10, 30]$ and $\textbf{w}_{\textbf{v},\boldsymbol{\omega}}\in[0.1, 0.3]$.
For testing, we set $\textbf{w}_{\textbf{p},\boldsymbol{\phi}}=[10, 20, 30]$, $\textbf{w}_{\textbf{v},\boldsymbol{\omega}}=[0.1,0.2,0.3]$.
For all cases, $\textbf{w}_\textbf{u}$ is fixed to $10^{-5}$.
Other system parameters are randomized as in Table \ref{table:DR}. 
At each setting, we run 10 simulations for 10 seconds each, and evaluate success rate and RMS velocity tracking error averaged over the entire $10\ s$ time window.

\begin{figure}[t]
    \centering
    \includegraphics[width=0.99\linewidth]{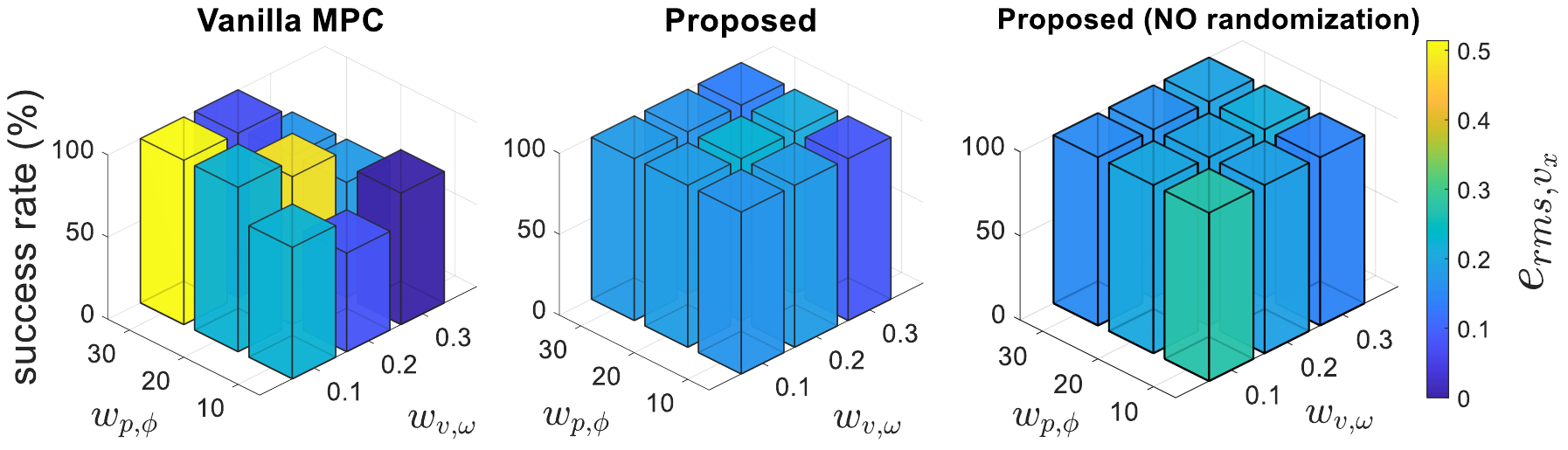}
    \caption{Results of control parameter robustness test. Each graph indicates success rate (height) and RMS error (color) of velocity tracking from corresponding parameter setup with (1) \textit{vanilla-MPC}, (2) proposed, and (3) proposed without randomization in simulation. Each weight label indicates that all corresponding weights (e.g., $\textbf{w}_{\textbf{p},\boldsymbol{\phi}}=[\textbf{w}_{\textbf{p}_{x,y,z}},\textbf{w}_{\boldsymbol{\phi}_{x,y,z}}]$) share the same value.}
    \label{fig:05} 
\end{figure}

As shown in Fig. \ref{fig:05}, \textit{vanilla-MPC} exhibits significant performance sensitivity to weights, while our method demonstrates consistent performance across all configurations. 
These results suggest that the residual modules effectively reduce dependency on manual tuning, allowing for successful task execution under a broader set of parameter settings.

In addition, we conducted an experiment without weight randomization during training. 
In this case, weights are fixed to $\textbf{w}_{\textbf{p},\boldsymbol{\phi}}=10$, $\textbf{w}_{\textbf{v},\boldsymbol{\omega}}=0.3$.
Fig. \ref{fig:05} indicates that performance remained relatively consistent even without randomization. 

\subsection{Consistent Performance via MPC}

Conventional end-to-end RL methods often struggle to handle situation that deviate from their training distribution \cite{kumar2021rmarapidmotoradaptation, 9001157}. 
In contrast, many model-based controllers are capable of producing consistent motions across a wide range of states.
We aim to combine the consistency of MBA with the robustness of RL-based methods, and validate whether our proposed framework offers improved consistency compared to existing approaches.

For comparison, we design an end-to-end RL controller (\textit{baseline-RL}) as a baseline \cite{ji2022concurrent}. 
This framework concurrently learns a state estimator to predict indirect states, such as translational velocity and contact probability, alongside the control policy.
The policy is trained using PPO under same conditions and network structure in Table \ref{table:PPO}.

The baseline is then compared with our method. 
For both cases, the robot is trained in the domain as shown in Table \ref{table:DR}. 
Note that the payload is only given on robot's trunk in simulation.
In the experiment, a $3\ kg$ payload is attached to both front and rear legs on the right side, creating an OOD situation from simulation. 
The robot is commanded to follow the given velocity command while maintaining its posture. 

\begin{figure}[t]
    \centering
    \includegraphics[width=0.99\linewidth]{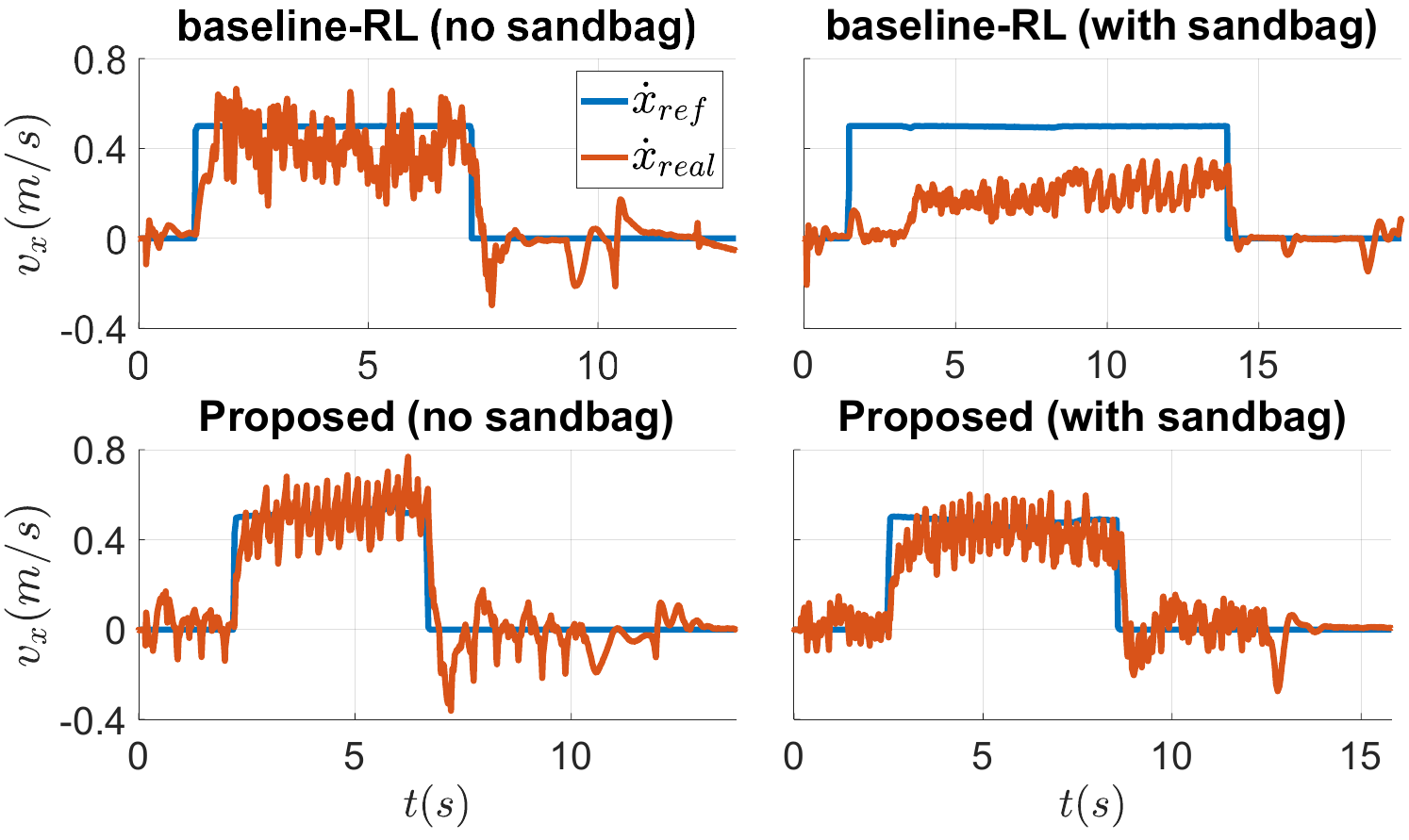}
    \caption{An experimental result of out of distribution test.}
    \label{fig:06}
\end{figure}

The experimental results are shown in Fig. \ref{fig:06}. Without any payload, both \textit{baseline-RL} and \textit{res-all} successfully track the given velocity command. 
However, the velocity tracking performance of \textit{baseline-RL} drops significantly with external payloads.
In contrast, the proposed method maintains performance without notable degradation.
These results indicate that integrating MBAs with residual modules enhances robustness and enables reliable motion, even in the presence of uncertainties out of training domain.

\subsection{Comparative Analysis with Baselines}

Our proposed methodology integrates different types of residual modules into the nominal controller. 
To analyze each module's contribution to performance, we conduct an ablation study with baseline methods.

In simulations, a robot is commanded to traverse terrain with varying roughness, procedurally generated using Perlin noise.
The roughness level $h_{env}$ is defined as the maximum height difference of the terrain. 
In each simulation, the robot is commanded to follow the given command $v^*_x=1.0\ m/s$, $\omega^*_z=1.0\ rad/s$. 
Each training is conducted with the setup in Table \ref{table:DR}, except for $M_{payload}=[0.9, 1.8]M$, $h_{env}=[0, 0.2]\ m$.
The test environment consists of $h_{env}=0.2\ m$ and $M_{payload}=[1.75, 2, 2.25]M$, respectively.
We set success rate and RMS of velocity tracking error as criteria, and conduct 100 simulations for each combination of environment condition and controller.

For the ablation study, we evaluate a total of 8 controller configurations, each representing a different combination of the residual modules.
For instance, \textit{fpos-phase} only compensates footstep position and phase.
The baselines include methods from Chen et al. \cite{10610453}, which use RL to find auxiliary actions in the joint space (\textit{jpos}) and dynamics space (\textit{dynRL}); 
Yang et al. \cite{pmlr-v164-yang22d}, who proposed a hierarchical framework with a learned gait transition module (\textit{phase}); 
and the online adaptation strategy from \cite{9525285}, which regresses model discrepancies within a sliding window (\textit{resdyn-window}).

As summarized in Table \ref{table:ablation}, our method consistently achieves a high success rate and stable tracking performance across different environmental conditions, while most baseline methods exhibit greater fluctuations in performance depending on the level of uncertainty. 

The analysis revealed specific weaknesses in baselines. 
For example, the success rate of \textit{fpos-phase} drops sharply under large uncertainties, highlighting the necessity of dynamics compensation. 
While \textit{fpos-phase-dynRL} shows performance comparable to our method, our framework exhibits improved learning efficiency, which will be discussed later. 
Conversely, \textit{baseline-RL} adopt an overly conservative strategy, refusing to follow commands under high uncertainty to avoid falling. 
Furthermore, the poor performance of \textit{phase} and \textit{resdyn-window} highlights that compensating for inaccuracies in both footsteps and dynamics is crucial for robustness.

\begin{figure}[t]
    \centering
    \includegraphics[width=0.95\linewidth]{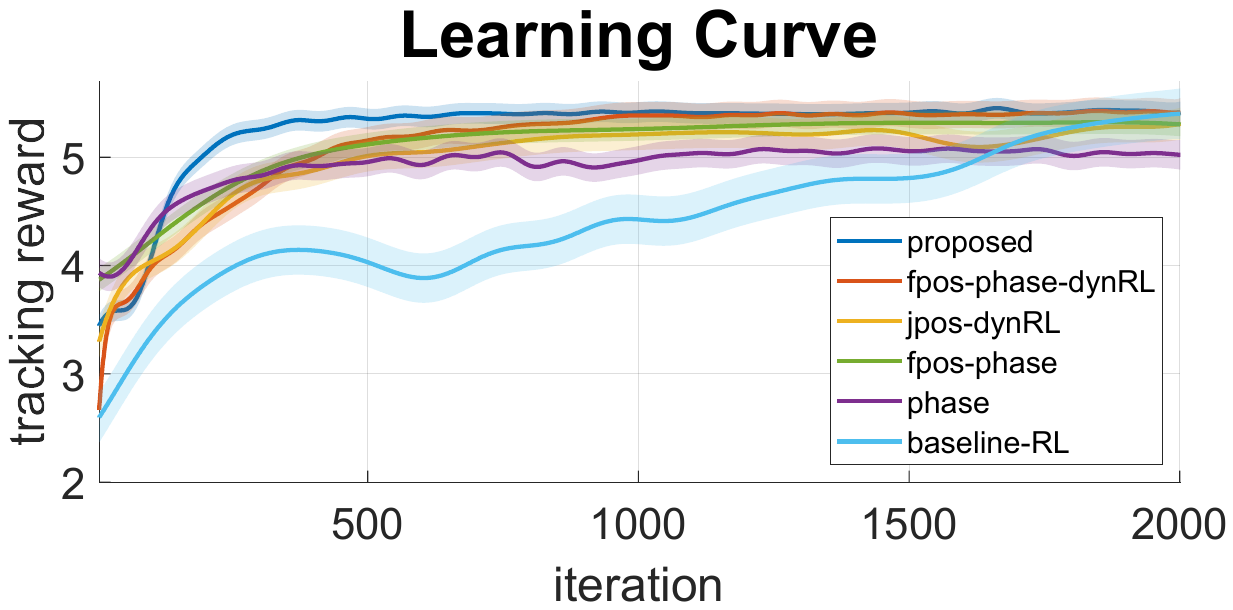}
    \caption{Comparative results of learning efficiency.}
    \label{fig:07}
\end{figure}

Lastly, we examine the impact of each residual module on learning efficiency during training. 
As depicted in Fig. \ref{fig:07}, our proposed method not only achieves the highest tracking reward but also demonstrates the stable and fastest convergence. 
This suggests that the proper combination of residual modules with nominal MBA leads to more refined locomotion, enabling more sample-efficient learning. 

\begin{table}[t]
\centering
\noindent
\caption{\textbf{Result of Comparative Analysis}}
{\scriptsize 
\begin{tabular}{N G G G}\toprule
\multicolumn{1}{c}{\textbf{Name}} &
    \multicolumn{1}{c}{\tiny$M_p=1.75M$} &
    \multicolumn{1}{c}{\tiny$M_p=2M$} &
    \multicolumn{1}{c}{\tiny$M_p=2.25M$} \\
\cmidrule(lr){2-4}
\multicolumn{1}{c}{res-all (proposed)} & 
    \multicolumn{1}{c}{0.195 (76\%)} &
    \multicolumn{1}{c}{{\textbf{0.185}} {\textbf{(74\%)}}} &
    \multicolumn{1}{c}{{\textbf{0.174}} {\textbf{(58\%)}}} \\
\multicolumn{1}{c}{fpos-phase-dynRL} &
    \multicolumn{1}{c}{0.200 {\textbf{(88\%)}}} &
    \multicolumn{1}{c}{0.207 (72\%)} &
    \multicolumn{1}{c}{0.196 (57\%)} \\
\multicolumn{1}{c}{jpos-dynRL} &
    \multicolumn{1}{c}{{\textbf{0.185}} (64\%)} &
    \multicolumn{1}{c}{0.189 (55\%)} &
    \multicolumn{1}{c}{0.194 (3\%)} \\
\multicolumn{1}{c}{fpos-phase} &
    \multicolumn{1}{c}{0.201 (71\%)} &
    \multicolumn{1}{c}{0.220 (72\%)} &
    \multicolumn{1}{c}{- (0\%)} \\
\multicolumn{1}{c}{phase} &
    \multicolumn{1}{c}{0.277 (39\%)} &
    \multicolumn{1}{c}{0.306 (11\%)} &
    \multicolumn{1}{c}{0.312 (1\%)} \\
\multicolumn{1}{c}{baseline-RL} &
    \multicolumn{1}{c}{0.805 (59\%)} &
    \multicolumn{1}{c}{0.766 (40\%)} &
    \multicolumn{1}{c}{0.800 (28\%)} \\
\multicolumn{1}{c}{vanilla-MPC} &
    \multicolumn{1}{c}{0.314 (41\%)} &
    \multicolumn{1}{c}{0.320 (25\%)} &
    \multicolumn{1}{c}{- (0\%)} \\
\multicolumn{1}{c}{resdyn-window} &
    \multicolumn{1}{c}{0.309 (39\%)} &
    \multicolumn{1}{c}{0.293 (30\%)} &
    \multicolumn{1}{c}{0.282 (14\%)} \\
\bottomrule
\end{tabular}
\caption*{\raggedleft\scriptsize 
Values indicate 'RMS velocity error (success rate)'.

Lowest RMSs and highest success rates are denoted in bold.}
\label{table:ablation}
}
\end{table}

%%%%%%%%%%%%%%%%%%%%%%%%%%%%%%%%%%%%%%%%%%%%%%%%%%%%%%%%%%%%%%%%%%%%%%%%%%%%%%%%
\section{CONCLUSIONS} \label{ch:05}

In summary, we propose a novel hybrid framework that combines the strengths of both MBAs and LBAs. The residual modules are designed to compensate for the limitations of heuristics in each part of the conventional MBA. The benefits of the proposed framework are validated through both simulations and hardware experiments.

We observe that our framework can produce more adaptive locomotion while still providing consistent performance, even beyond the training domain. Additionally, our approach can alleviate the hyperparameter sensitivity of nominal MBA. 
% Furthermore, the proper combination and design of residual modules can facilitate efficient training.
Furthermore, our method demonstrates improved learning efficiency compared to the baselines.

However, to fully validate the general applicability of our method, broader evaluations are necessary under various conditions. 
We expect that with further module design and training in more diverse domains, the proposed framework can be extended to handle a wider range of uncertainties, for example, uncertain terrain properties such as softness, compliance, and slipperiness.

\ifCLASSOPTIONcaptionsoff
  \newpage
\fi



%\bibliographystyle{IEEEtran}
\bibliographystyle{ieeetr}
\bibliography{reference}


%\vfill


\end{document}